\documentclass{article}

\usepackage{graphicx}
\usepackage{amsmath} 
 \usepackage[preprint]{neurips_2025}

\usepackage[utf8]{inputenc} % allow utf-8 input
\usepackage[T1]{fontenc}    % use 8-bit T1 fonts
\usepackage{hyperref}       % hyperlinks
\usepackage{url}            % simple URL typesetting
\usepackage{booktabs}       % professional-quality tables
\usepackage{amsfonts}       % blackboard math symbols
\usepackage{nicefrac}       % compact symbols for 1/2, etc.
\usepackage{microtype}      % microtypography
\usepackage{xcolor}         % colors

\title{Exploring Diffusion Models for
Generative\\ Forecasting of Financial Charts}

\author{%
  Taegyeong Lee$^{1,2}$,
  Jiwon Park$^{2}$,
  Kyunga Bang$^{2}$,
  Seunghyun Hwang$^{1,2,3}$,
  Ung-Jin Jang$^{1}$\thanks{Corresponding author} \\
  \\
  $^{1}$FnGuide Inc. \\
  $^{2}$GenAI in Finance, MODULABS \\
  $^{3}$Department of Applied Data Science, Sungkyunkwan University \\
  \\
  \texttt{taegyeonglee@fnguide.com},  
  \texttt{mary000605@ewha.ac.kr},  \\
  \texttt{\{kbang1002, hsh1030\}@g.skku.edu, } 
  \texttt{coorung77@fnguide.com} \\
}

\begin{document}

\maketitle

\begin{abstract}
Recent advances in generative models have enabled significant progress in tasks such as generating and editing images from text, as well as creating videos from text prompts, and these methods are being applied across various fields. However, in the financial domain, there may still be a reliance on time-series data and a continued focus on transformer models, rather than on diverse applications of generative models. In this paper, we propose a novel approach that leverages text-to-image model by treating time-series data as a single image pattern, thereby enabling the prediction of stock price trends. Unlike prior methods that focus on learning and classifying chart patterns using architectures such as ResNet or ViT, we experiment with generating the next chart image from the current chart image and an instruction prompt using diffusion models. Furthermore, we introduce a simple method for evaluating the generated chart image against ground truth image. We highlight the potential of leveraging text-to-image generative models in the financial domain, and our findings motivate further research to address the current limitations and expand their applicability.
\end{abstract}

\section{Introduction}
Recent advances in Large Language Models (LLMs)\cite{chen2024internvl, achiam2023gpt,team2023gemini} and diffusion-based generative models~\cite{brooks2022instructpix2pix,rombach2022high,kawar2022imagic,lee2023generating} have enabled powerful cross-modal capabilities such as generating and editing images from text. In finance, however, most stock price prediction~\cite{lu2025enhancing,xiao2024comparative} still relies on single-modality time-series models like LSTMs~\cite{graves2012long} or Transformers~\cite{vaswani2017attention}, which may struggle to capture visually discernible signals (chart patterns, candlestick shapes)~\cite{velay2018stock,lin2021improving,zhu2024pmanet} and to integrate heterogeneous factors such as news or sentiment. Existing image-based studies~\cite{velay2018stock, pizon2025image,bang2023cnn} typically perform classification with architectures like ResNet~\cite{he2016deep} or ViT~\cite{dosovitskiy2020image}, without explicitly modeling the temporal evolution of chart patterns and relying only on chart snapshots without integrating signals such as volume or technical indicators.

In real markets, human traders often interpret price movements through visual patterns—head-and-shoulders, double bottoms, candlestick wicks~—because these configurations implicitly encode market psychology such as fear, greed, and indecision~\cite{martinssson2017short,mersal2025enhancing}. Traditional time-series approaches, while effective at modeling sequential dependencies, lose the structural “shape” context of price movements, struggle to jointly represent multiple heterogeneous signals (e.g., price, volume, indicators)~\cite{liang2022stock}, and offer limited interpretability from a trader’s perspective.

In this work, we explore a novel approach that generates the next chart image from the current chart and an instruction prompt using a text-to-image diffusion model~\cite{rombach2022high}. To support this, we construct paired datasets of input charts, instruction prompts, and edited chart images, and fine-tune a U-Net in latent space. Because the generated outputs are stochastic and differ from time-series predictions, we further introduce a simple image-marking evaluation that compares generated images to ground-truth charts via RGB analysis. 

We conducted preliminary experiments on cryptocurrency data, and the results suggest that our method is feasible and promising, though performance remains limited. Our method represents an exploratory step toward leveraging generative models in financial forecasting, and suggests potential extensions such as incorporating sentiment signals from news articles.

Our main contributions are as follows:
\begin{itemize}
\item We propose a novel approach that leverages a text-to-image generative model, treating time-series financial data as visual patterns for next-step chart generation.
\item We introduce a paired dataset construction method combining chart images, instruction prompts, and edited images, enabling the model to learn both technical indicators (RSI, MACD) and visual chart patterns.
\item We introduce a simple image-marking evaluation method to quantify prediction accuracy from generated images. Also we show experiment results on cryptocurrency data, and discuss the limitations and opportunities for extending this approach to richer multi-modal signals such as sentiment.
\end{itemize}

\section{Method}
\vspace{-0.2cm}
\begin{figure*}[t] %%%
\begin{center}
\includegraphics[width=1.0\textwidth]{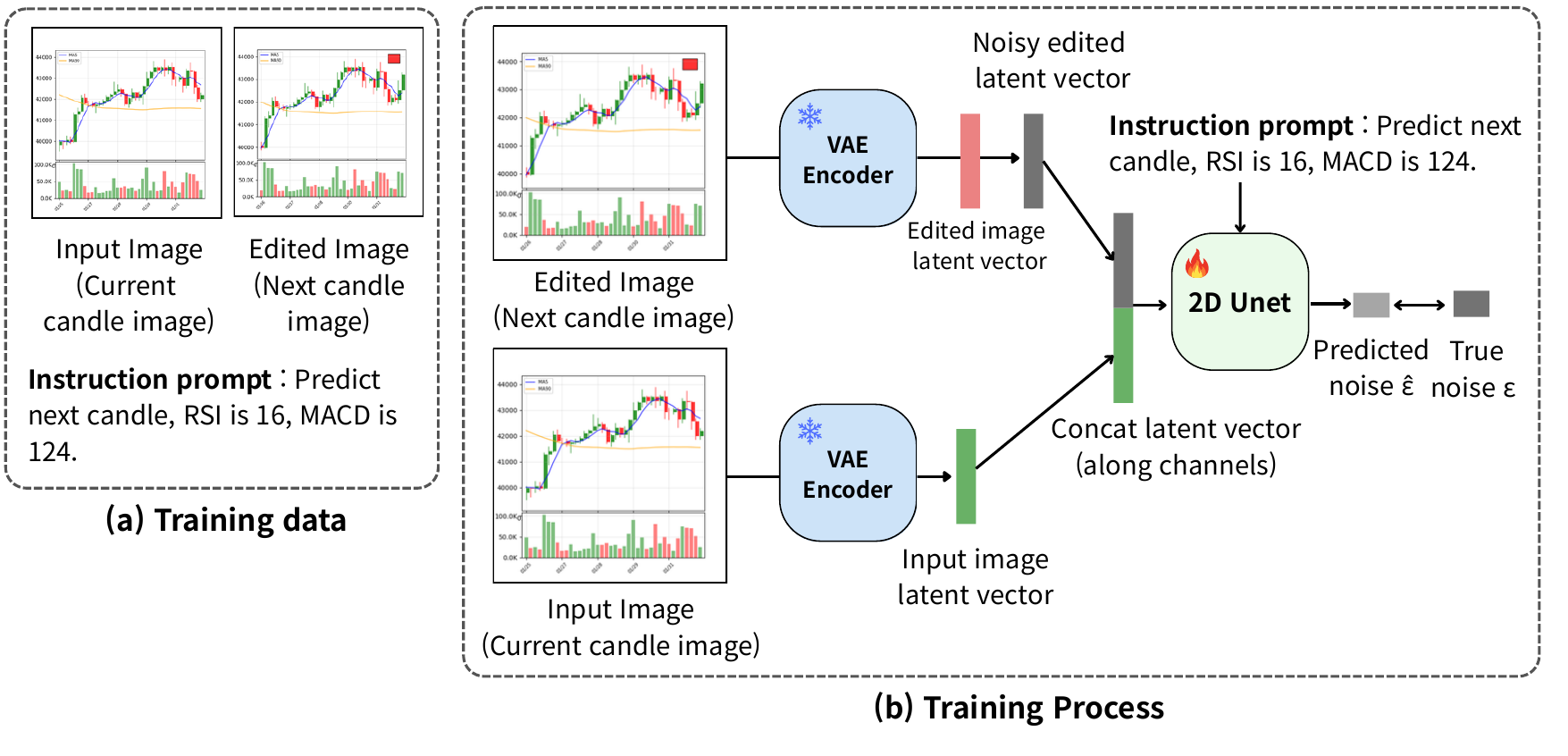}
\end{center}
\vspace{-0.4cm}
\caption{\textbf{Overall architecture of our method}. \textbf{(a)} Training data: We construct a paired dataset consisting of an input image, an edited image, and an instruction prompt. \textbf{(b)} Training Process : We encode the input image and the edited image, and then fine-tune the 2D U-Net of Stable Diffusion with the instruction prompt.
}
\label{figure_overview}
\vspace{-0.4cm}
\end{figure*} %%% Figure

\subsection{Training Dataset}
\vspace{-0.2cm}
\label{sec:training_dataset}
 Inspired by InstructPix2Pix~\cite{brooks2022instructpix2pix}, we propose a simple yet effective method for fine-tuning Stable Diffusion~\cite{rombach2022high} to learn chart patterns and generate next-candle charts based on given instructions. As shown in Figure~\ref{figure_overview}~(a), we construct paired datasets consisting of an input image, an edited image, and an instruction prompt. In our dataset, the input image is represented as a 4-hour candlestick chart at the ${n}$-th timestep, incorporating trading volume as well as the SMA5 and SMA90 lines. This allows the model to learn not only from the candlestick chart itself, but also from the relationships with trading volume and moving averages. The edited image is constructed as the candlestick chart at timestep ${n}{+}3$, with an evaluation marker placed at the upper-right corner of the image. A red mark is assigned if the price increases by more than 2\%, a blue mark if it decreases, and a black mark otherwise. The instruction prompt consists of the RSI and MACD values at timestep ${n}$, formatted with the prefix: \texttt{Predict next candle}.

\begin{table*}[t]
\begin{minipage}{0.67\linewidth}
\centering
\small
\caption{Category-wise classification performance. We generate our images for the test input image, and then evaluate our model by analyzing the RGB values to determine its consistency with the ground truth.}
\vspace{0.2cm}
\begin{tabular}{l@{\hspace{0.1cm}}cccc}
\toprule
Category & Precision (\%) & Recall (\%) & F1-Score (\%) & Support \\
\midrule
blue     & 8.06  & 11.63 & 9.52  & 43  \\
red      & 9.26  & 14.93 & 11.43 & 67  \\
black    & 85.60 & 77.94 & 81.59 & 671 \\
\midrule
Overall Acc. & \multicolumn{4}{c}{68.89\% (538/781)} \\
\bottomrule
\end{tabular}
\label{tab:performance}
\end{minipage}
\hfill
\begin{minipage}{0.32\linewidth}
\centering
\small
\caption{Confusion matrix of classification results. P. blue, P. red, and P. black indicate the predicted classes, while A. blue, A. red, and A. black indicate the actual classes.}
\vspace{0.2cm}
\begin{tabular}{l@{\hspace{0.1cm}}c@{\hspace{0.1cm}}c@{\hspace{0.1cm}}c}
\toprule
 & P.\ blue & P.\ red & P.\ black \\
\midrule
A. blue  & 5  & 4  & 34  \\
A. red   & 3  & 10 & 54  \\
A. black & 54 & 94 & 523 \\
\bottomrule
\end{tabular}
\label{tab:confusion}
\end{minipage}
\vspace{-0.4cm}
\end{table*}

\subsection{Training Process}
As illustrated in Figure~\ref{figure_overview}~(b), we freeze the VAE of Stable Diffusion~1.5~\cite{rombach2022high} and fine-tune only the 2D U-Net. 
Given an input image $\mathbf{I}_{\text{in}}$ (at timestep $n$) and an edited image $\mathbf{I}_{\text{edit}}$ (at timestep $n{+}3$), 
the frozen VAE encoder produces latent vectors 
$\mathbf{l}_z, \mathbf{n}_e \in \mathbb{R}^{4 \times 64 \times 64}$, 
where $\mathbf{n}_e$ is further perturbed with gaussian noise. 
The two latents are concatenated along the channel dimension to form 
$\mathbf{x} \in \mathbb{R}^{8 \times 64 \times 64}$, 
and the U-Net input stem is modified accordingly. 

The instruction prompt provides the technical indicators at timestep $n$, formatted as:  
\texttt{Predict next candle, RSI is \{value\}, MACD is \{value\}},  
and is injected via cross-attention. 
The U-Net is trained to predict the noise $\hat{\boldsymbol{\epsilon}}_\theta$ from the noisy latent $\mathbf{x}_t$ and instruction $\mathbf{c}$, 
with the standard denoising objective:
\[
\mathcal{L}(\theta) = 
\mathbb{E}_{\mathbf{x}, \mathbf{c}, t, \boldsymbol{\epsilon}}
\big[
\| \boldsymbol{\epsilon} - 
\text{UNet}_\theta(\mathbf{x}_t, t, \mathbf{c}) \|_2^2
\big].
\]
Through this training process, our method leverages the image generation capability of Stable Diffusion~\cite{rombach2022high} to learn chart-specific patterns while incorporating instruction prompts, thereby enabling the generation of future chart images from the current chart.

\subsection{Inference for Next Chart Generation}
Since we fine-tune Stable Diffusion 1.5~\cite{rombach2022high}, the model preserves its image generation capability while learning chart-specific patterns. Given the current chart image with RSI and MACD values as an instruction prompt, the model can generate a candlestick chart four hours ahead. The generated chart includes the evaluation mark, trading volume, and moving averages, similar to the edited image, as illustrated in Figure~\ref{fig:samples}.

\section{Experiments}
\subsection{Experimental Setup}
\textbf{Dataset.} Our method represents time series data as image charts, making it possible to fine-tune for any asset. We focus particularly on Bitcoin price, as its data is easy to collect and its patterns are clearly visible. Following Section~\ref{sec:training_dataset}, we construct a paired dataset consisting of 2,419 training data pairs in total. We use 4-hour candlestick charts of Bitcoin future prices with Binance exchange from January 1, 2024, to March 1, 2025, as the input images. Each training image contains a total of 40 candlesticks. The edited image is constructed in the same manner as the input image, but with an evaluation marker placed at the upper-right corner. 

\textbf{Implementation detail.} 
We fine-tune Stable Diffusion 1.5~\cite{rombach2022high} using an NVIDIA A100 80GB GPU, with a batch size of 16, a gradient accumulation step of 4, and a total of 28,000 optimization steps. Our inference step is set to 20, image guidance scale to 1.0, and guidance scale to 2.0. We use the EulerAncestralDiscreteScheduler as the image scheduler.

\textbf{Evaluation.}
We construct an evaluation dataset using 4-hour candlestick charts of Bitcoin futures from the Binance exchange, covering the period from March 17, 2025 to July 31, 2025. Our evaluation dataset consists of three paired components: input image, instruction prompt, and edited image, for a total of 781 pairs. We simply classify the generated image by reading the RGB values of this mark region and mapping them to one of the three classes via a color-thresholding rule. Accuracy is computed as the fraction of samples where the predicted class matches the ground-truth class.

\begin{figure*}[t] %%%
\begin{center}
\includegraphics[width=1.0\textwidth]{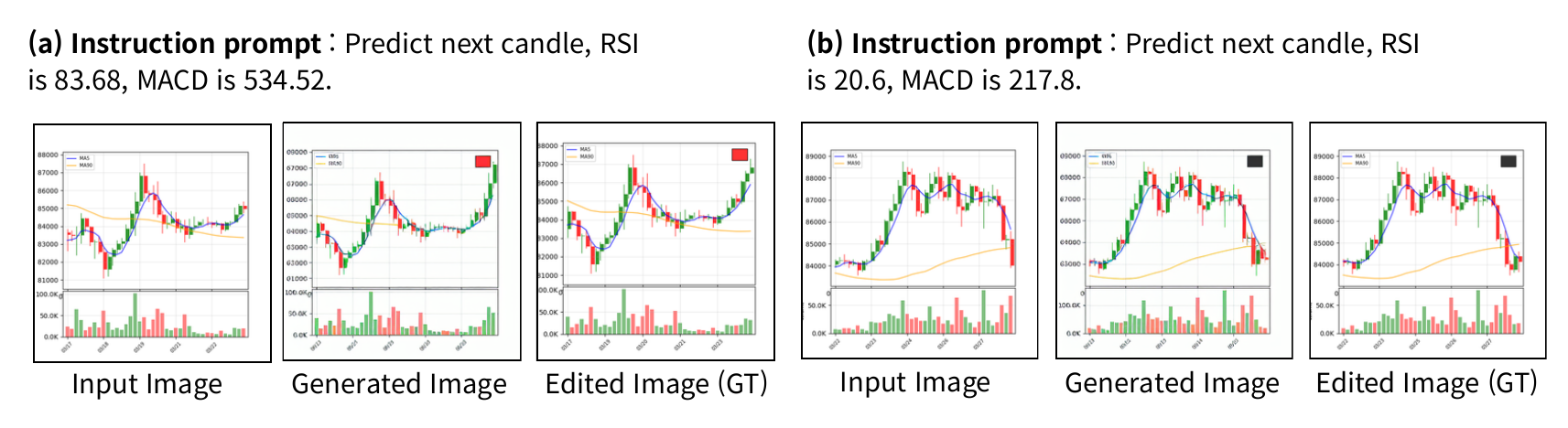}
\end{center}
\vspace{-0.2cm}
\caption{\textbf{Generated samples}. We generate the image from an input image and an instruction prompt, while the edited image serves as the ground truth. Our results demonstrate that the model is capable of sufficiently learning the visual patterns of charts.
}
\vspace{-0.4cm}
\label{fig:samples}
\end{figure*} %%% Figure

\subsection{Quantitative Results}
\vspace{-0.2cm}
Table~\ref{tab:performance} presents the classification results, with an overall accuracy of 68.89\% (538/781). The performance is mainly driven by the black class (F1-score 81.59\%), while the blue and red classes show much lower F1-scores (9.52\% and 11.43\%), as confirmed by the confusion matrix in Table~\ref{tab:confusion}. Many blue and red samples are misclassified as black, indicating that minority-class patterns are less distinctive and easily absorbed into the majority category.

These findings suggest that the proposed generative approach may capture broad trend patterns, while it appears less effective in distinguishing finer variations across underrepresented classes. Contributing factors include dataset imbalance, the difficulty of diffusion models in retaining chart-specific structures, and imperfect alignment between generated and ground truth images.

Although the results leave room for improvement, our method is intended as an exploratory step. Our experiments show that text-to-image generative models can encode meaningful chart-like structures, and we believe that methods such as class balancing, domain-aware conditioning, or hybrid generative–discriminative frameworks could improve quantitative performance in future work.

\subsection{Qualitative Results}
\vspace{-0.2cm}

As shown in Figure~\ref{fig:samples}, we use the input image and instruction prompt as inputs. Based on these, our model generates the next candlestick chart, which can be seen in the generated image. The generated image shares similarities with the edited image, and we can observe that it is generated by taking into account trading volume, moving averages, as well as RSI and MACD indicators. Despite its limited quantitative performance, the model can generate chart patterns and diverse samples to visualize and anticipate scenarios, opening possibilities for generative forecasting charts. We expect traders could use it as a scenario simulation tool to visually explore market psychology.
\vspace{-0.2cm}
\section{Limitation and Conclusion}
\vspace{-0.2cm}
\textbf{Limitation.} Our approach still suffers from  limitation. The RGB-based evaluation remains simplistic and may overlook the true financial validity of generated charts. The dataset size is limited and centered on cryptocurrency, restricting broader applicability. In addition, the predictive performance of our model is still modest, highlighting that current generative approaches are not yet competitive with traditional forecasting methods. Nevertheless, we believe that future work holds significant opportunities for improvement. For example, instruction prompts could be enriched with diverse external signals, such as financial news, FOMC announcements, or market sentiment reports. Multi chart inputs (e.g., combining 15-minute and 4-hour candlesticks) could provide the model with richer temporal context. More powerful generative backbones, such as Stable Diffusion XL~\cite{podell2023sdxl}, may improve both fidelity and consistency. Beyond this, domain-specific evaluation metrics, larger and more diverse datasets, and multimodal integration with order book depth or social media sentiment could substantially enhance both accuracy and interpretability.

In conclusion, while our current model shows limited predictive performance, this work illustrates a novel direction: reframing time-series forecasting in finance through the lens of generative models. By treating financial data as visual patterns for chart generation, we open up a research pathway that blends generative modeling with technical and multimodal analysis, paving the way for more advanced and practically useful financial forecasting systems.

\subsubsection*{Acknowledgments}
This research was supported by Brian Impact Foundation, a non-profit organization dedicated to the advancement of science and technology for all.

{\small
\bibliographystyle{plain}
\bibliography{ref}
}

\end{document}